\documentclass[10pt,journal,compsoc]{IEEEtran}
\usepackage{amsmath}
\usepackage{amssymb}
\usepackage{mathtools}
\usepackage{multirow}
\usepackage[dvipsnames]{xcolor}
\usepackage[algo2e]{algorithm2e} 
\usepackage{algorithm}

\DeclareMathOperator*{\argmin}{argmin}
\usepackage{algorithmic}
\usepackage{multirow}
\usepackage{tabularx}
\usepackage{comment}
\usepackage{subfig}
\usepackage{float}
\newcommand{\Lagr}{\mathcal{L}}
\newcommand{\Dcal}{\mathcal{D}}
\newcommand{\Scal}{\mathcal{S}}
\newcommand{\Wcal}{\mathcal{W}}

\ifCLASSOPTIONcompsoc
  \usepackage[nocompress]{cite}
\else
  \usepackage{cite}
\fi
\ifCLASSINFOpdf
\else
\fi
\begin{document}
%
\title{Pred\&Guide: Labeled Target Class Prediction for Guiding Semi-Supervised Domain Adaptation}
%
%
%
%

\author{Megh~Bhalerao, Anurag~Singh,
        and Soma Biswas,~\IEEEmembership{Senior Member, IEEE}
\IEEEcompsocitemizethanks{\IEEEcompsocthanksitem Megh Bhalerao, Anurag Singh and Soma Biswas collaborated in this work at the Department of Electrical Engineering at the Indian Institute of Science, Bangalore, India. \protect
E-mail (corresponding author): somabiswas@iisc.ac.in
}
}

\IEEEtitleabstractindextext{%
\begin{abstract}
Semi-supervised domain adaptation aims to classify data belonging to a target domain by utilizing a related label-rich source domain and very few labeled examples of the target domain.
Here, we propose a novel framework, Pred\&Guide, which leverages the inconsistency between the predicted and the actual class labels of the few labeled target examples to effectively guide the domain adaptation in a semi-supervised setting.
Pred\&Guide consists of three stages, as follows 
(1) First, in order to treat all the target samples equally, we perform unsupervised domain adaptation coupled with self-training; 
(2) Second is the label prediction stage, where the current model is used to predict the labels of the few labeled target examples, and 
(3) Finally, the correctness of the label predictions are used to effectively weigh source examples class-wise to better guide the domain adaptation process.
Extensive experiments show that the proposed Pred\&Guide framework achieves state-of-the-art results for two large-scale benchmark datasets, namely Office-Home and DomainNet. 
\end{abstract}

}

\maketitle

\IEEEdisplaynontitleabstractindextext

%
\IEEEpeerreviewmaketitle

\IEEEraisesectionheading{\section{Introduction}\label{sec:introduction}}

%
%
%
%
\IEEEPARstart{I}{n} real world, the training data distribution is often different from the testing data distribution, necessitating that deep learning models learnt using data from one domain are adapted to data from a different distribution. 
Unsupervised Domain Adaptation (UDA)~\cite{ganin2016domainadversarial,ganin2015dann,li2020unsupervised, 6247911} addresses this problem by utilizing a related label rich source domain and unlabeled samples from the target domain.
Recent research has shown that only unlabeled target samples may not be sufficient to counter the domain shift, and thus the area of Semi-supervised Domain Adaptation (SSDA)~\cite{saito2019semisupervised,saito2018adversarial,6618936,kim2020attract,ijcai2020-130,7298826} is gaining traction, where additionally, a few labeled targets are utilized to aid the adaptation process. 

The advantage of SSDA over UDA approaches lies in the manner in which the few labeled targets are utilized in the adaptation process.
A straightforward approach is to include them along with the labeled source examples in the Cross-Entropy loss ~\cite{saito2019semisupervised,saito2018adversarial}. 
Though this approach helps in improving the final performance, it does not fully leverage the usefulness of the labeled target samples, since they form a very small part of the final loss as compared to the large number of source examples.

In this work, we propose a novel framework {\bf Pred\&Guide}, which utilizes the class label prediction inconsistencies on the few target labeled examples to effectively guide the domain adaptation process.
Pred\&Guide has three main stages:
(1) \textit{UDA with Self-Training}: Access to a few labeled target data can create bias in the domain adaptation process from the very beginning towards those samples as also noted in~\cite{li2020unsupervised,long2017unsupervised}.
Thus, we propose to utilize an UDA framework to start the domain adaptation process in an unbiased manner, which utilizes all the target data, but ignores the available labels of the few target examples.
Simultaneously, we also compute the pseudo-labels of all the unlabeled data using the current model, and perform strong augmentations of those target examples, for which we are confident about their pseudo-labels  ~\cite{sohn2020fixmatch,berthelot2019mixmatch,berthelot2019mixmatch}.
(2) \textit{Labeled Target Prediction}: The model trained in the first stage is used to predict the labels of the targets (which have ground truth labels available) to analyze the adaptation process, and further guide it.
(3) \textit{Source Example Weighting}: We surmise that different source examples either help or hinder the process of domain adaptation. 
Based on the inconsistency of the predicted and true labels of the few labeled target examples,
Pred\&Guide steers the adaptation process by appropriately weighing the source samples according to their cosine similarity to the labeled target examples of the corresponding class.
Specifically, the source samples in the neighborhood of the incorrectly classified targets are given more weight and the ones which are far away are down-weighed, following a linear weighing scheme as described in Section \ref{source example weighing}, so that the classifier can eventually classify the labeled targets correctly.

We extensively evaluate the proposed Pred\&Guide framework on two large-scale benchmark datasets, namely  Office-Home~\cite{officehome} and DomainNet~\cite{domainnet}.
Pred\&Guide outperforms all the state-of-the-art approaches for both the datasets.
The main contributions of our work are as follows: 
\begin{enumerate}
\item We propose a novel approach for semi-supervised domain adaptation, termed \textit{Pred\&Guide}, by leveraging the prediction inconsistency of the labeled targets. 
\item We propose to effectively weigh the source examples based on the label prediction inconsistency to guide the DA process. 
\item Extensive experiments on two benchmark datasets show the effectiveness of the proposed framework.
\end{enumerate}
We now describe the related work in literature followed by the proposed approach and results of extensive evaluation.

\section{Related Work}
\label{rw}
Here, we provide pointers to the related work in literature on domain adaptation, data augmentation and semi-supervised learning. \\
{\bf Domain Adaptation:}
Most of the current literature in domain adaptation can be broadly categorized into unsupervised (UDA) and semi-supervised domain adaptation (SSDA). 
UDA approaches, which use only unlabeled target domain data have been explored rigorously in \cite{ganin2016domainadversarial,ganin2015dann,long2017unsupervised,Saito_2018_CVPR,pmlr-v28-gong13}. Initial approaches used statistical methods such as Maximum Mean Discrepancy \cite{Sejdinovic_2013}, Correlation Alignment (CORAL) \cite{sun2016correlation} and Geodesic Flow Kernel \cite{6247911} to handle the mismatch in feature distribution. 
More recently, DL based UDA approaches such as \cite{ganin2015dann,ganin2016domainadversarial} aim to learn a domain invariant features by a domain confusion module for feature alignment. It has been observed that UDA performance can be boosted significantly with just a few labeled examples from the target domain, as noted in works such as \cite{saito2019semisupervised,kim2020attract,ijcai2020-130,nam2020reducing}. 
Most of the recent SSDA approaches use $1-3$ labeled target examples per class.
The seminal work in~\cite{saito2019semisupervised} uses a mini-max entropy to compute domain-invariant class-representatives. 
In \cite{li2021cross} authors propose an adversarial loss instead of min-max entropy.
In~\cite{ijcai2020-130}, adversarial examples are used to bridge the source and target domain gap. 
~\cite{nam2020reducing, singh2021clda} use
a style-agnostic network and contrastive learning for reducing the source and target domain gaps.

However, majority of approaches treat examples of labeled target no differently than the labeled source, thus not fully leveraging information contained in them. Some works have attempted to use labeled target in triplet loss for bringing source closed to target~\cite{li2021ecacl} and~\cite{liang2021domain} propose an effective pseudo labeling strategy using labeled target. 
Presence of few labeled target examples in SSDA setting results in intra-domain discrepancy, which is recently noted and addressed in~\cite{kim2020attract}.
We also address this critical issue Pred\&Guide.
\\~\\
{\bf Data Augmentation:}
Data augmentation~\cite{cubuk2019randaugment,cubuk2019autoaugment,tanaka2019data} is widely used as a regularizer to prevent overfitting. 
Different data augmentation policies have been proposed in works such as ~\cite{cubuk2019autoaugment,cubuk2019randaugment,sohn2020fixmatch}.
From a pool of augmentations, ~\cite{cubuk2019randaugment} randomly selects augmentation to be used. 
~\cite{cubuk2019autoaugment} selects an augmentation during model training using reinforcement learning.
Other approaches such as CT-Augment have been proposed in~\cite{sohn2020fixmatch,berthelot2020remixmatch} which uses a similar strategy as AutoAugment \cite{cubuk2019autoaugment}, but uses a fixed algorithm for assigning augmentation magnitudes.
In our work, we use an augmentation based consistency-regularization, based on RandAugment~\cite{cubuk2019randaugment} to assign pseudo-labels to the unlabeled targets to boost the performance.
\\~\\
{\bf Semi-Supervised Learning (SSL):}
SSL approaches utilize a small amount of labeled data and a huge amount of unlabeled data for the required task~\cite{chen2020selftraining,10.5555/2976040.2976107,article,sohn2020fixmatch,miyato2018virtual,tarvainen2018mean,berthelot2019mixmatch,berthelot2020remixmatch}, which reduces the labeling cost.
Several SSL algorithms use consistency constraint, i.e. they aim to minimize the entropy between prediction of different versions of a data-point~\cite{sohn2020fixmatch,berthelot2019mixmatch,berthelot2019mixmatch,tarvainen2018mean}.

Mean-Techer~\cite{tarvainen2018mean} trains a student and a teacher model and forces their predictions to match. 
The student model is updated at every iteration, while the teacher model is updated with a suitable momentum. 
Virtual Adversarial Training~\cite{miyato2018virtual} uses adversarial perturbations to train the SSL model. 
Pseudo-labeling~\cite{article} predicts labels of the unlabeled data and uses these as true labels for supervised learning. Recent approaches such as Mixmatch~\cite{berthelot2019mixmatch}, ReMixmatch~\cite{berthelot2020remixmatch} and Fixmatch~\cite{sohn2020fixmatch} provide an elegant framework for SSL, leveraging consistency regularization and MixUp ~\cite{zhang2018mixup} to obtain state-of-the-art results for SSL. 
The proposed SSDA framework Pred\&Guide is inspired by these seminal SSL approaches, but utilizes the few labeled target domain examples in a novel manner to guide the adaptation process.

\section{Problem Definition}
We first discuss the problem statement and the notations used.
In this work, we address the SSDA task, where there are very few labeled target examples available (i.e. one and three shot settings).
The training data consists of (1) \textit{labeled source}: $\mathcal{D}_{ls} = \{x^s_i,y^s_i\}_{i=1}^{i=n_s}$, where $n_s$ is the number of labeled source examples, (2) \textit{unlabeled target}: $\Dcal_{ut} = \{x^{ut}_i\}_{i=1}^{i=n_{ut}}$, where $n_{ut}$ is the number of unlabeled target examples and (3) \textit{few labeled target}: $\mathcal{D}_{lt} = \{x^{t}_i,y^t_i\}_{i=1}^{i=n_t}$, where $n_t$ is the number of labeled target examples. 
Here  $x_i$ denotes the $i^{th}$ data-point and $y_i$ denotes the corresponding class label, which belongs to one of the $K$ classes (same for both source and target domains). 
We test the model's performance on $\Dcal_{ut}$. 
For all the experiments, we have a feature extractor network $F$ (\textit{with parameters} $\Theta_F$), sequentially followed by a classifier $C$ (\textit{with parameters} $\Theta_C$). 
\section{Proposed Pred\&Guide Framework}
\begin{figure*}[ht]
\centering
\includegraphics[width=0.7\textwidth]{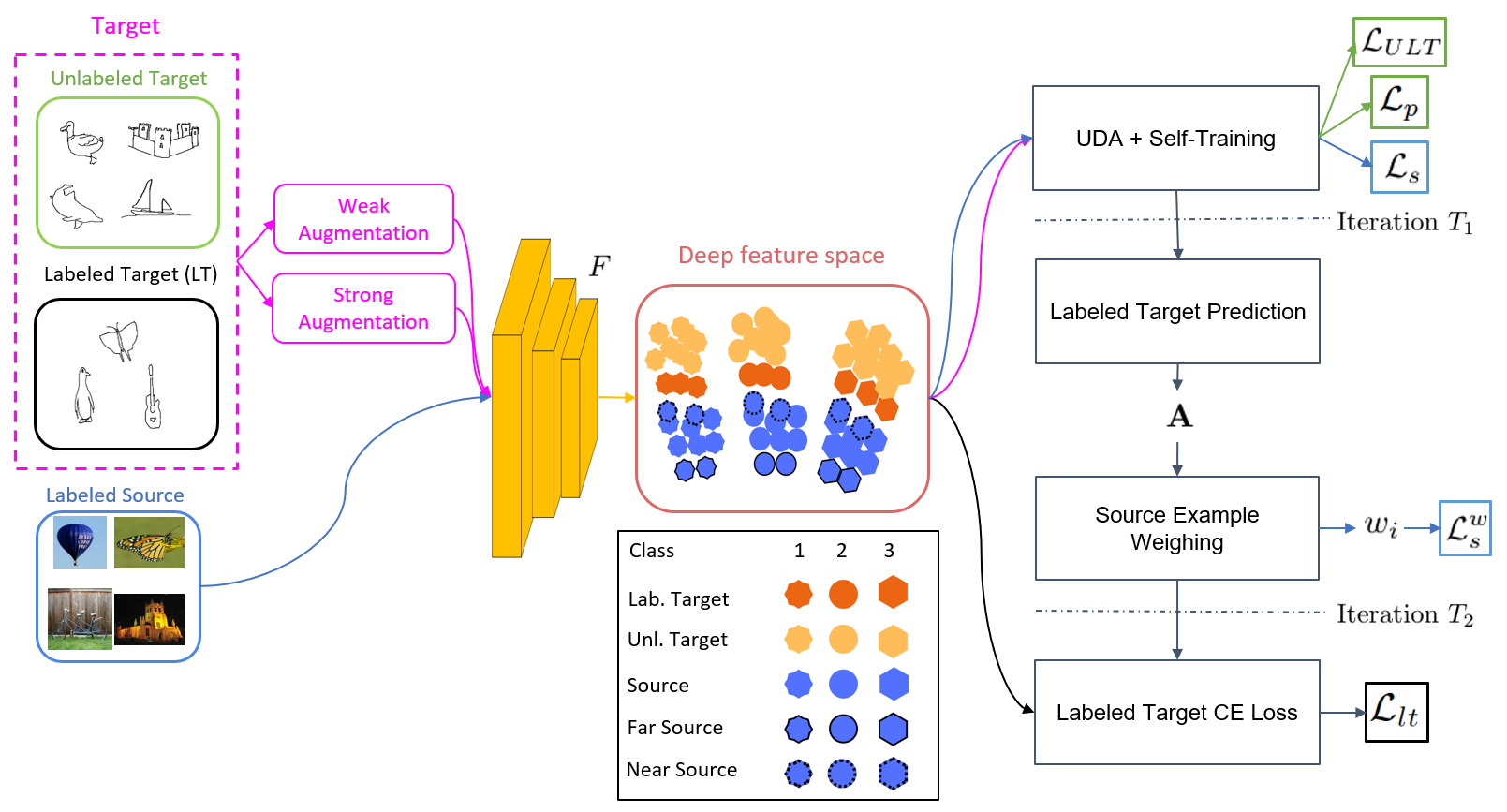}
\caption{Figure showing the complete Pred\&Guide framework and its different modules. 
The different modules get activated at different iteration steps, when the corresponding losses are computed as seen from the indicator functions in equation (\ref{l1 equation}). 
The distance to source samples are calculated efficiently using a feature bank (\textit{not shown in figure}) in the feature space.\color{black} }
\label{fig:main_flowchart}
\end{figure*}
The proposed SSDA framework has three sequential modules, namely (1) unsupervised domain adaptation (UDA) with self-training (2) labeled target prediction, where we predict the labels of the target domain data using the current model, for which we have the ground truth labels and finally (3) source sample weighting, where the source examples are effectively weighted using the prediction inconsistencies of the labeled target examples in the previous step. Figure \ref{fig:main_flowchart} illustrates the proposed Pred\&Guide framework and its different modules.
Each of the modules with its motivation are explained below in detail. 
\subsection{UDA with Self-Training}
In SSDA, though few labeled target samples are available in the training data, using them from the beginning may bias the domain adaptation process towards these examples, as noted in recent works~\cite{li2020unsupervised,long2017unsupervised}.
This has motivated us to use an UDA approach to initiate the adaptation process in an unbiased manner, which treats all the target samples equally. 
For this, any UDA approach can be utilized. 
Most UDA approaches usually learn a domain invariant feature extractor and a classifier~\cite{ganin2015dann,ganin2016domainadversarial,li2020unsupervised}. 
Let us denote this model as $\mathbf{U_M}$. 

Inspired by the recent works in consistency regularization and self-training/pseudo-labeling~\cite{sohn2020fixmatch,berthelot2020remixmatch,berthelot2019mixmatch}, while training $\mathbf{U_M}$, we incorporate an augmentation-consistency based self-training to further aid the unsupervised learning, which is briefly described below for completion.

During every training iteration, for all the unlabeled target data $x^{ut}\: \in \: \Dcal_{ut}$ we compute
\begin{equation}
\label{eq:strong and weak augmentations}
\vspace{-8px}
    x^{ut_s} = \Scal(x^{ut}); \qquad
    x^{ut_w} = \Wcal(x^{ut})   
\end{equation}
where $x^{ut_s}$ and $x^{ut_w}$ are the strong and weak augmentations of the given data $x^{ut}$.
The strong augmentation function $\Scal$ is obtained using RandAugment \cite{cubuk2019randaugment}, while the weak augmentation function $\Wcal$ is simply flipping, random cropping and padding.
In every forward pass of the current model $\mathbf{U_M}$, for all the unlabeled target samples, we compute the predicted class probability distribution of the weak and strong augmentations, denoted by $p^{ut_w}$ and $p^{ut_s}$ respectively. 
Using $p^{ut_w}$, we compute the one-hot pseudo-label
\begin{equation}
    y^{ut_w} = \mathbf{O}(\arg \max(p^{ut_w}))
\end{equation}
where the function $\mathbf{O}$ produces a valid one-hot vector corresponding to the predicted label.
Our per example pseudo-labeled target loss is:
\begin{equation}
\label{pseudo label loss}
\begin{split}
    \Lagr_p = \mathbf{1}(\max(p^{ut_w})>\tau)\,\mathbf{H}(y^{ut_w},p^{ut_s})
\end{split}
\end{equation}
Here, we only consider the loss for the unlabeled targets whose prediction confidence is greater than a threshold $\tau$. 
$\mathbf{H}(.,.)$ is the standard cross-entropy function, where the first parameter is the ground truth label and the second is the predicted class probability distribution over the labels. 
The proposed method calculates the pseudo-labels on-the-fly for every example as opposed to static pseudo-labels being updated at different time-steps~\cite{article}.

For the source, we consider standard cross entropy loss $\mathbf{H(.,.)}$, until $t < T_{1}$. 
Here, $T_{1}$ is the number of iterations for which UDA with self-training runs till the validation accuracy saturates.    
After $T_1$, we use the weighted cross entropy for the source examples, as explained later in Section \ref{source example weighing}. 
The loss for source example $(x^s, y^s)$ is given as:
\begin{equation}
\label{source standard CE loss}
    \Lagr_s = \mathbf{H}(y^s,p^s)
\end{equation}
where $y^s$ is the label of the source example and $p^s$ is the predicted class probability distribution of $x^s$.  

For the unlabeled target example $x^{ut_w}$ we use the mini-max entropy approach~\cite{saito2019semisupervised}, with the loss function denoted by $\Lagr_{ULT}$, where $\Lagr_{ULT}$ is given by: 
\begin{equation}
\vspace{-3px}
\label{unlabeled target loss}
    \Lagr_{ULT} = -\sum_{k=1}^{K}(p^{ut_w}_{k})\log(p^{ut_w}_{k})
\end{equation}
Thus, in this step, the domain adaptation uses all the target samples, but without the labels of the few target examples which are available.
In the next stage, we check how good the current model is in predicting the labels of the labeled targets and its augmentations, which is then used to guide the adaptation process.
      
\subsection{Labeled Target Prediction}

In this stage, we leverage the additional label information provided for few target examples to analyze how well the current model has adapted to the target domain. 
Since the number of labeled targets is very less, we leverage their strong and weak augmentations denoted by $x^{t_s}$ and $x^{t_w}$ for this analysis, calculated following equation (\ref{eq:strong and weak augmentations}). 
Let us denote this new augmented data as $\Dcal_{lt_{a}}$. 


First, the current model is used to predict the class labels of the examples in $\Dcal_{lt_{a}}$ for which we know the ground truth class labels.
Computing the label prediction accuracy for each class in this manner gives a weak indication of the class-wise adaptation of the current model.
For example, if most of the (labeled) target samples of a class have been classified correctly, it indicates that domain adaptation is successful for that class, otherwise, it is not satisfactory.
Obviously, as the number of labeled target samples increases, this becomes a strong indicator of the class-wise domain adaptation accuracy, hence we use the labeled data along with its augmentations. 
We define the class-wise accuracy vector as follows:
\begin{equation}
    \label{accuracy on labeled target}
    \mathbf{A} = [a_1,a_2,\dots,a_K],\;\; \mathbf{A} \in \mathbb{R}^{K \times 1}
\end{equation}
and $a_i$'s are the individual class accuracies as calculated on $\Dcal_{lt_{a}}$ using the current  model $ \mathbf{U_M}$, and $K$ is the total number of classes.
Pred\&Guide aims to utilize the class-wise information about adaptability, rather than just the domain-wise adaptability, since different classes vary in their ability to adapt. 
This follows from the fact that it may easier for some classes of a domain to adapt (if those classes \textit{look} similar across domains), while it may be difficult for others.

The goal is to aid the classes which have not yet adapted satisfactorily, based on the accuracy computed.
For the classes which have lesser accuracy, more (less) weightage is given to the neighboring (far) source samples of the corresponding labeled target examples.
This helps the corresponding class prototypes adapt to the target domain samples, as will be explained in detailed next. 

\subsection{Source Example Weighing}
\label{source example weighing}
Once UDA is performed, the label prediction in the previous module is used to weigh source examples accordingly at regular intervals. 
If a target example is wrongly classified, we weigh its neighbouring source samples of the same class relatively more (based on a linear weighing scheme as explained next), which will guide the model towards classifying this target sample correctly.
To further aid the adaptation, we also relatively down-weigh the source samples of the same class far away from the wrongly classified labeled targets, since these source examples hinder the adaptation process.
To perform this step efficiently with reduced computational complexity, we use a feature bank to store the representations of the source examples as described next. \\ \\
\textbf{Feature Bank Based Source Identification:} To compute the distances of source examples of the same class as the labeled target examples, we need to compute the distance between the labeled target and the source examples of the same class. 
To compute this distance efficiently, we maintain a feature bank $\mathbf{S}$ as defined below:
\begin{equation}
\vspace{-7px}
    \mathbf{S} = [s_1,s_2,\dots,s_{n_s}],\;\;\mathbf{S} \in \mathbb{R}^{d_f \times n_s}
\end{equation}
where $s_i$ is the representation of the $i^{th}$ source example. $n_s$ denotes the number of examples in the source domain (as defined in Section 3) and
$d_f$ is the dimension of the representation space. $\mathbf{S}$ is updated \textit{on the fly} batch-wise with momentum $m_s$ as:
\begin{equation}
\label{eq:7}
\vspace{-5px}
    \mathbf{S}_{t + 1} \leftarrow m_s\mathbf{S}_{t} + (1-m_s)\,\mathbf{f}_{bs}
\end{equation}
where $\mathbf{f}_{bs}$ are the representations of the current batch of source examples. 
$\mathbf{S}_t$ denotes the feature bank at a given iteration-step $t$. 
For simplicity we assume that in~(\ref{eq:7}), only the source examples corresponding to the ones in the mini-batch are updated. 
Now, we describe how our linear weighing scheme works.

\begin{figure*}[ht!]
\centering
\subfloat{\includegraphics[width=0.3\textwidth]{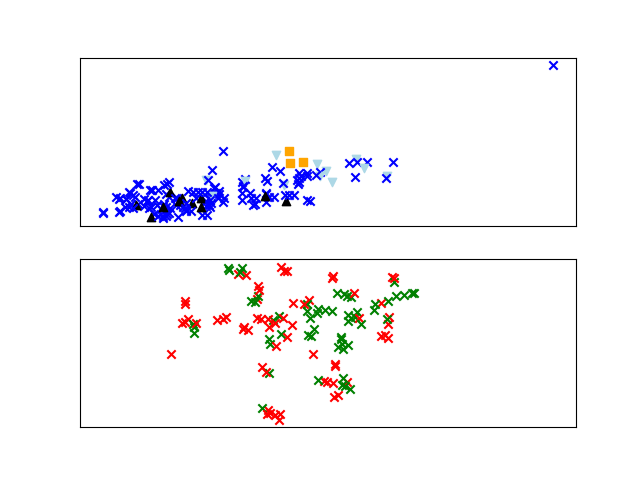}}
\subfloat{\includegraphics[width=0.3\textwidth]{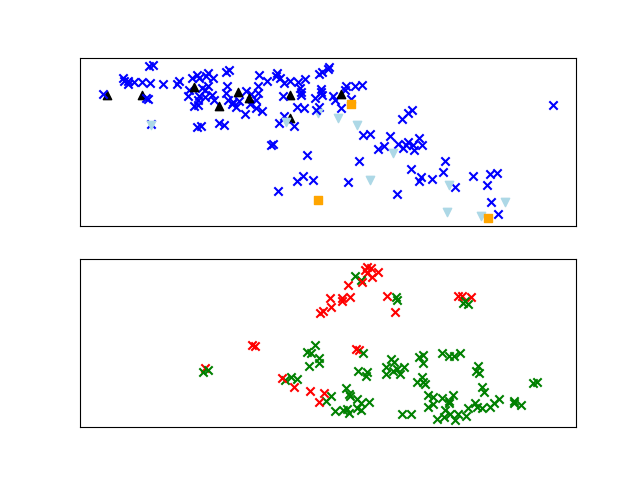}}
\subfloat{\includegraphics[width=0.3\textwidth]{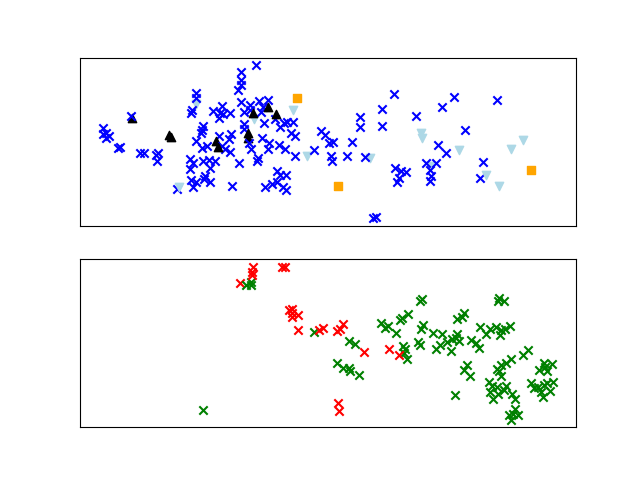}}
\caption{Figure showing the t-SNE ~\cite{JMLR:v9:vandermaaten08a} plots at different stages of training for \textbf{one} particular class in the P to C domain adaptation setting for DomainNet. 
The labeled source and labeled target (above) and the unlabeled target (below) are shown in separate plots for clarity (but share the same set of axes). 
The target classification accuracy for this class at these three stages (from left to right) are $47.9\%$, $76.7\%$ and $81.5\%$ for the $2000^{th}$, $7500^{th}$, and $15000^{th}$ iterations respectively, as can be clearly seen from the increasing number of correctly classified target examples. \color{black} 
Color coding - 
\textcolor{blue}{$\times$}: unweighted labeled source,
\textcolor{cyan}{$\blacktriangledown$} and 
\textcolor{black}{$\blacktriangle$
}: 5 near and 5 far weighted samples respectively; 
\textcolor{orange}{Orange Square}: labeled targets; 
\textcolor{red}{$\times$}: wrongly classified unlabeled targets; \textcolor{OliveGreen}{$\times$}: correctly classified unlabeled targets.}
\label{fig:tsne}
\end{figure*}

Let $\mathcal{D}^{k}_{ls} \subset \mathcal{D}_{ls}$
denote the set of source examples belonging to class $k\:\in\:\{1,2,\dots,K\}$. 
Similarly, let $\mathcal{D}^k_{lt} \subset \mathcal{D}_{lt}$
be the set of labeled targets belonging to class $k$. 
Now for every element $x^{t,k}$ in $\mathcal{D}_{lt}^{k}$ (sample index $i$ is omitted for simplicity), we compute an ordered set
$\Tilde{\mathcal{D}}_{ls}^k=\{x^{s,k}_j\in \mathcal{D}_{ls}^{k}\text{ s.t. } d(s^{s,k}_{j+1}, F(x^{t,k})) > d(s^{s,k}_{j},F(x^{t,k}))\}$ where $s^{s,k}_{j}$ is the corresponding feature from the feature bank $\mathbf{S}_t$ for sample $x^{s,k}_{j}$ and $d(.,.)$ is cosine similarity between the feature representations of the datapoints.
Let $min_{sim}$ and $max_{sim}$ denote the minimum and maximum cosine similarities from the set of source examples ($\Tilde{\mathcal{D}}^{k}_{ls}$) of the given class $k$. We compute the class wise max and min weights as:
\begin{equation}
\label{eq: min and max weight}
\begin{split}
        max_{w} = 1 + \phi/\exp(a_k) \\
        min_{w} = 1 - \phi/\exp(a_k)
\end{split}
\end{equation}
where $a_k$ is the accuracy computed using the predictions of the labeled targets of class $k$, as computed in the previous step of Pred\&Guide and $\phi$ is a hyper-parameter. Thus, the $i^{th}$ source sample belonging to class $k$ is weighted with $w_i$ as
    \begin{equation}
    \label{calculation of source example weights}
        w_i = m \times (d(s^{s,k}_{i},F(x^{t,k})) - min_{sim}) + min_w
    \end{equation}
where $m$ is the slope of the linear weighting scheme
\begin{equation}
    m = \frac{max_w - min_w}{max_{sim} - min_{sim}}
\end{equation}
In other words, we up-weigh the near source samples and down-weigh the far source samples, with the weight varying according to a linear function dependent on the similarity of the source sample with respect to the labeled target sample as in equation (\ref{calculation of source example weights}).
\color{black}
If the accuracy for class $k$ is less (i.e. $a_k$ is small), the prototype for that particular class probably has not adapted properly.
For these classes, the weights of source examples $w_i$ in the neighbourhood of the labeled targets of that class is increased, and those which are far away are decreased, which will help the domain adaptation process.
Thus, the weighted loss for a source example $(x^s,y^s)$ with predicted class probability distribution $p^s$ is expressed as follows:
\begin{equation}
\label{weighted source loss}
    \Lagr^{w}_{s} = w^s \times \mathbf{H}(y^s,p^s)
\end{equation}
where $w^s$ is weight of the example whose loss is being calculated. Figure \ref{fig:tsne} illustrates the near and far examples at different iterations of Pred\&Guide using t-SNE plots~\cite{JMLR:v9:vandermaaten08a}. 

\begin{figure*}[t!]
\centering
\includegraphics[width=0.9\textwidth,height=180px]{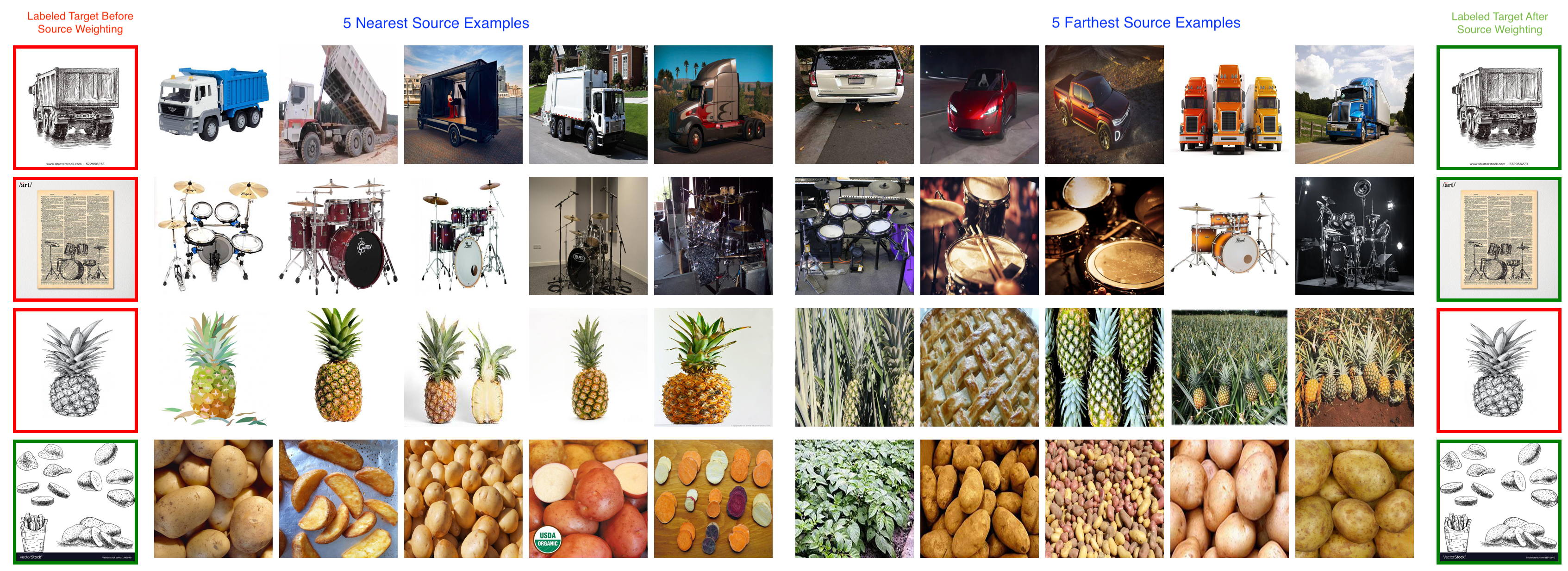}
\caption{Figure showing 5 nearest and farthest source samples corresponding to labeled targets of four different classes.
Green (red) boundaries of the labeled targets indicate whether they are correctly (incorrectly) classified after UDA with self-training (first column) and after Pred\&Guide training is complete with effective source weighting (last column). 
We observe that for the first two rows, Pred\&Guide is able to effectively correct the labels, while in the third row, though the near and far source examples are quite intuitive, the final prediction is still incorrect. The last row shows one example which is always correctly classified.
This example is for the Real to Sketch setting of DomainNet with ResNet34 backbone. }
\label{fig:near_far_examples}
\end{figure*}

\color{black}
Once we have used the labeled target accuracy to calculate and assign weights for the source examples $T_1$ number of iterations as described above, we bring in the labeled target examples $(x^{t},y^{t})$, with predicted class probability $p^{t}$, into the training.
For these samples, we use the standard cross-entropy loss starting at iteration $T_{2}$ (number of iterations after which source example weighing validation accuracy saturates) as follows: 
\begin{equation}
\label{labeled target CE loss}
    \Lagr_{lt} = \mathbf{H}(y^{t},p^{t})
\end{equation}

\begin{algorithm}[h]
\DontPrintSemicolon
{\small
  \textbf{Input:}  Feature extractor $F$, classifier $C$, total number of iterations $T$  \\
  \KwData{Labeled source $\Dcal_{ls}$, unlabeled target $\Dcal_{ut}$, labeled target $\Dcal_{lt}$, augmented labeled target $\Dcal_{lt_{a}}$.}
  \While{$t < T$}{
  Calculate pseudo-labeled target loss $\Lagr_p$ according to (\ref{pseudo label loss}).\\
  Calculate unlabeled target loss $\Lagr_{ULT}$ according to (\ref{unlabeled target loss}).\\
  \If{$t > T_1$}
    {
        \If{$t \bmod T_n == 0$}{
        Calculate accuracy on $\Dcal_{lt_a}$ to obtain $\mathbf{A}$ as in equation (\ref{accuracy on labeled target}).\\
        Calculate and update source example weights using equation (\ref{calculation of source example weights}). 
        }
    Calculate and backpropagate $\Lagr^{w}_s$ according to equation (\ref{weighted source loss}).
    }
    \ElseIf{$t < T_1$}
    {
    Calculate cross-entropy loss for the labeled source examples $\Lagr_s$ according to (\ref{source standard CE loss}).
    }
    \If{$t > T_2$}{
    Calculate cross-entropy loss for the labeled target examples $\Lagr_{lt}$ according to (\ref{labeled target CE loss}).
    }
    $t \leftarrow t + 1$
  }
  }

\caption{{\small Proposed Pred\&Guide algorithm}}
\label{alg1}
\end{algorithm}
\subsection{Complete Pred\&Guide Framework}
The complete Pred\&Guide Framework is summarized in this section. 
As stated earlier, we optimize the parameters of the feature extractor $\Theta_F$ and classifier $\Theta_C$ using the mini-max approach~\cite{saito2019semisupervised}:
\begin{equation}
\label{minimization equation}
\begin{split}
    \hat{\Theta}_F = \underset{\Theta_F}{\argmin} \; \mathbf{\Lagr_1} + \lambda \Lagr_{ULT} \\
    \hat{\Theta}_C = \underset{\Theta_C}{\argmin} \; \mathbf{\Lagr_1} - \lambda \Lagr_{ULT} 
\end{split}
\end{equation}
Here $\mathbf{\Lagr_1}$ is the sum of all losses except $\Lagr_{ULT}$ loss given by
\begin{equation}
\label{l1 equation}
    \begin{split}
    \mathbf{\Lagr_{1}} = \Lagr_p  
    + \mathbf{1}(t < T_{1}) \Lagr_s 
    + \mathbf{1}(t > T_{1}) \Lagr^{w}_{s} + 
    \mathbf{1}(t > T_{2})\Lagr_{lt}
    \end{split}
\end{equation}
Here, $T_{1}$ is the iteration after which we start weighing the source examples and $T_{2}$ is the iteration at which the labels of the few target examples are exposed to the training algorithm.
More details about calculating $T_1$ and $T_2$ are given in Section \ref{impl details}. \color{black}
The complete Pred\&Guide algorithm is given in Algorithm~\ref{alg1}.
\section{Experimental Evaluation}
\label{experimental evaluation}
Here, we provide details of the extensive experiments conducted to evaluate the efficacy of Pred\&Guide for semi-supervised domain adaptation. 
First, we discuss the benchmark datasets used, the implementation details before reporting the results. 
We also conduct detailed analysis of Pred\&Guide and perform ablation studies to evaluate the usefulness of the individual modules. \\ \\ 
{\bf Datasets Used and Evaluation Protocol: }
To evaluate Pred\&Guide, we use two large-scale benchmark domain adaptation datasets, \textbf{DomainNet}~\cite{domainnet} and \textbf{Office-Home}~\cite{officehome}. DomainNet consists of $6$ domains with $345$ classes, comprising of about $0.6M$ images. Since the entire DomainNet dataset is noisy, we select $4$ domains (\textit{Real, Painting, Clipart, Sketch}) and $126$ classes with $7$ standard domain adaptation scenarios, which are generally used to benchmark DA methods ~\cite{saito2019semisupervised,ijcai2020-130,kim2020attract,saito2018adversarial}.
In addition to DomainNet, we also evaluate Pred\&Guide on the Office-Home dataset, which is a smaller, but challenging dataset. It consists of $65$ classes and $4$ domains (\textit{Real, Clipart, Product, Art}). We evaluate Pred\&Guide on all the $12$ possible adaptation scenarios for Office-Home. \\ \\
{\bf Implementation Details:}
\label{impl details}
The code is implemented using PyTorch ~\cite{paszke2019pytorch} on a single Nvidia RTX-2080 GPU. 
The underlying domain adaptation technique that we used in Pred\&Guide is the very successful mini-max entropy approach~\cite{saito2019semisupervised}.
We use Resnet34 \cite{he2015deep} and AlexNet \cite{NIPS2012_c399862d} for the feature extractor network $F$ to compare with state-of-the-art approaches. 
For the classifier $C$, we use a $ 4096 \times K$ fully-connected layer for AlexNet and a $512 \times K$ fully-connected layer for ResNet34.
We use SGD optimizer, with the starting learning rate of $0.01$, momentum of $0.9$, weight decay of $0.0005$ and $\lambda = 0.1$ for the weight of $\Lagr_{ULT}$. 
We use batch size of $24$ and $32$ for Resnet34 and Alexnet backbones respectively on labeled samples, and twice the batch size for unlabeled data. We set $\phi = 0.5$ in Equation (\ref{eq: min and max weight}) for DomainNet and $\phi=0.1$ for Office-Home, and $m_s = 0.1$ in Equation \ref{eq:7} for all our experiments.
We run the first module, i.e. UDA with self-training for $T_{1}$ iterations. $T_{1}$ is the iteration-step when UDA with self-training has converged. 
Then, the weights of the source examples are computed every $T_n$ iterations. $T_n = 1000$ for DomainNet and $T_n = 140$ for Office-Home.
Finally, we bring in the labeled target examples in the CE loss at the $T_{2}$ iteration. $T_{2}$ is started when the source example weighing performance has converged.
To calculate $T_1$ and $T_2$, we consider the validation accuracy to be converged when it has not increased for $500$ iterations. \\ \\ 
{\bf Evaluation for SSDA: }
The results of Pred\&Guide for Office-Home dataset with AlexNet backbone for both $1$-shot and $3$-shot scenarios are reported in Table~\ref{office-home-table}.
Comparison with the recent approaches show that Pred\&Guide achieves the state-of-art results on an average for both the settings and also outperforms the others for most of the individual domain pair settings.
The results on DomainNet dataset for both $1$ and $3$ shots (i.e. one and three target samples of each class in the target domain are labeled) using AlexNet and ResNet34 backbones are reported in Table~\ref{Visdatable}. 
The results of all the other approaches has been directly  taken from ~\cite{saito2019semisupervised}.
We observe that the proposed approach significantly outperforms all the recent SSDA approaches, thus justifying its usefulness.

\begin{table*}[ht]
\centering
\scalebox{0.7}{
\begin{tabular}{|c|c|c|c|c|c|c|c|c|c|c|c|c|c|}
\hline
Method& R to C & R to P & R to A & P to R  & P to C & P to A & A to P & A to C & A to R & C to R & C to A & C to P & Mean\\
\hline
\multicolumn{14}{c}{One-Shot}\\
\hline 
S+T~\cite{ranjan2017l2constrained} & 37.5& 63.1& 44.8& 54.3& 31.7& 31.5& 48.8& 31.1& 53.3& 48.5& 33.9& 50.8& 44.1\\
DANN~\cite{ganin2015dann}& 42.5& 64.2& 45.1& 56.4& 36.6& 32.7& 43.5& 34.4& 51.9& 51.0& 33.8& 49.4& 45.1\\
ADR~\cite{saito2018adversarial} & 37.8& 63.5& 45.4& 53.5& 32.5& 32.2& 49.5& 31.8& 53.4& 49.7& 34.2& 50.4& 44.5\\
CDAN~\cite{long2018conditional}& 36.1& 62.3& 42.2& 52.7& 28.0& 27.8& 48.7& 28.0& 51.3& 41.0& 26.8& 49.9& 41.2\\
ENT~\cite{10.5555/2976040.2976107} & 26.8& 65.8& 45.8& 56.3& 23.5& 21.9& 47.4& 22.1& 53.4& 30.8& 18.1& 53.6& 38.8\\
BiAT~\cite{ijcai2020-130}&-&-&-&-&-&-&-&-&-&-&-&-&49.6\\ 
MME~\cite{saito2019semisupervised} & 42.0& 69.6& 48.3& 58.7& 37.8& 34.9 & 52.5& \textbf{36.4}& \textbf{57.0}& \textbf{54.1}& \textbf{39.5}& 59.1& 49.2\\ 
{\bf Pred\&Guide} & \textbf{44.4} & \textbf{73.2} & \textbf{50.0} & \textbf{59.6} & \textbf{38.2} & \textbf{37.0} & \textbf{54.4} & 34.8 & 55.6 & 52.8 & 38.0 & \textbf{59.2} & \textbf{ 49.8 }\\
\hline
\multicolumn{14}{c}{Three-Shot}\\
\hline
S+T~\cite{ranjan2017l2constrained} & 44.6& 66.7& 47.7& 57.8& 44.4& 36.1& 57.6& 38.8& 57.0& 54.3& 37.5& 57.9& 50.0\\
DANN~\cite{ganin2015dann}& 47.2& 66.7& 46.6& 58.1& 44.4& 36.1& 57.2& 39.8& 56.6& 54.3& 38.6& 57.9& 50.3\\
ADR~\cite{saito2018adversarial} & 45.0& 69.3& 46.9& 57.3& 38.9& 36.3& 57.5& 40.0& 57.8& 53.4& 37.3& 57.7& 49.5\\
CDAN~\cite{long2018conditional}& 41.8& 69.9& 43.2& 53.6& 35.8& 32.0& 56.3& 34.5& 53.5& 49.3& 27.9& 56.2& 46.2\\
ENT~\cite{10.5555/2976040.2976107} & 44.9& 70.4& 47.1& 60.3& 41.2& 34.6& 60.7& 37.8& 60.5& 58.0& 31.8& 63.4& 50.9\\
BiAT~\cite{ijcai2020-130}&-&-&-&-&-&-&-&-&-&-&-&-&56.4\\
MME~\cite{saito2019semisupervised} & 51.2& 73.0& 50.3& 61.6& 47.2& 40.7& 63.9& 43.8& 61.4& 59.9& 44.7& 64.7& 55.2\\
APE~\cite{kim2020attract}& 51.9& 74.6& 51.2& 61.6& 47.9& 42.1& 65.5 & 44.5 & \textbf{60.9}& 58.1& \textbf{44.3}& 64.8& 55.6\\
{\bf Pred\&Guide} & \textbf{53.4} & \textbf{75.0} & \textbf{51.9} & \textbf{64.0} & \textbf{48.7} & \textbf{43.6} & \textbf{65.7} & \textbf{45.7} & 60.6 & \textbf{60.0} & 43.0 & \textbf{67.5} & \textbf{56.6}\\
\hline 
\end{tabular}
}
\caption{Semi-supervised DA results (\%) on Office-Home data for both $1$ and $3$-shot protocols using Alexnet backbone for four domains with $12$ total domain combinations.}
\label{office-home-table}
\end{table*}

\begin{table*}[ht]
    \centering
    \scalebox{0.7}{
    \begin{tabular}{|c|c|c|c|c|c|c|c|c|c|c|c|c|c|c|c|c|c|c|}
    \hline
         \multirow{2}{*}{Net} & \multirow{2}{*}{Method} & \multicolumn{2}{c}{R to C} & \multicolumn{2}{c}{R to P} & \multicolumn{2}{c}{P to C} & \multicolumn{2}{c}{C to S} & \multicolumn{2}{c}{S to P} & \multicolumn{2}{c}{R to S} & \multicolumn{2}{c}{P to R} & \multicolumn{2}{|c|}{Mean}\\
         & & \multicolumn{1}{c}{1 shot} & \multicolumn{1}{c}{3 shot} & \multicolumn{1}{c}{1 shot} & \multicolumn{1}{c}{3 shot} & \multicolumn{1}{c}{1 shot} & \multicolumn{1}{c}{3 shot} & \multicolumn{1}{c}{1 shot} & \multicolumn{1}{c}{3 shot} & \multicolumn{1}{c}{1 shot} & \multicolumn{1}{c}{3 shot} & \multicolumn{1}{c}{1 shot} & \multicolumn{1}{c}{3 shot} & \multicolumn{1}{c}{1 shot} & \multicolumn{1}{c}{3 shot} & \multicolumn{1}{|c}{1 shot} &
         \multicolumn{1}{c|}{3 shot} \\
         \hline
         \multirow{8}{*}{Alexnet}
         &S+T~\cite{ranjan2017l2constrained}& 43.3&47.1&42.4&45.0&40.1&44.9&33.6&36.4&35.7&38.4&29.1&33.3&55.8&58.7&40.0&43.4\\
         &DANN~\cite{ganin2015dann}&43.3&46.1&41.6&43.8&39.1&41.0&35.9&36.5&36.9&38.9&32.5&33.4&53.6&57.3&40.4&42.4\\
         &ADR~\cite{saito2018adversarial}& 43.1&46.2&41.4&44.4&39.3&43.6&32.8&36.4&33.1&38.9&29.1&32.4&55.9&57.3&39.2&42.7\\
         &CDAN~\cite{long2018conditional}&46.3&46.8&45.7&45.0&38.3&42.3&27.5&29.5&30.2&33.7&28.8&31.3&56.7&58.7&39.1&41.0\\
         &ENT~\cite{10.5555/2976040.2976107}& 37.0&45.5&35.6&42.6&26.8&40.4&18.9&31.1&15.1&29.6&18.0&29.6&52.2&60.0&29.1&39.8\\
        
         &MME~\cite{saito2019semisupervised}& 48.9&55.6&48.0&49.0&46.7&51.7&36.3&39.4&39.4&43.0&33.3&37.9&56.8&60.7&44.2&48.2\\
         &BiAT~\cite{ijcai2020-130}& 54.2&\textbf{58.6}& 49.2& 50.6& 44.0& 52.0& 37.7& 41.9& \textbf{39.6}& 42.1& 37.2& 42.0& 56.9& 58.8& 45.5& 49.4\\
         & APE~\cite{kim2020attract}& 47.7 & 54.6 & 49.0& 50.5& 46.9 & 52.1 & 38.5 &42.6& 38.5& 42.2 & 33.8 &38.7& 57.5& 61.4 &44.6& 48.9\\
         & {\bf Pred\&Guide} & \textbf{54.5} & 57.3  & \textbf{54.4}  & \textbf{56.7} &\textbf{52.8}& \textbf{56.9} & \textbf{41.6} & \textbf{48.3}  & 36.9 & \textbf{44.9} &\textbf{38.4}&\textbf{46.9}&\textbf{61.9}&\textbf{65.4}&\textbf{48.6}& \textbf{53.8}\\
         \hline
         \hline
         \multirow{8}{*}{Resnet34}
         & S+T~\cite{ranjan2017l2constrained}& 55.6&60.0&60.6&62.2&56.8&59.4&50.8&55.0&56.0&59.5&46.3&50.1&71.8&73.9&56.9&60.0\\
         & DANN~\cite{ganin2015dann}&58.2&59.8&61.4&62.8&56.3&59.6&52.8&55.4&57.4&59.9&52.2&54.9&70.3&72.2&58.4&60.7\\
         & ADR~\cite{saito2018adversarial}& 57.1&60.7&61.3&61.9&57.0&60.7&51.0&54.4&56.0&59.9&49.0&51.1&72.0&74.2&57.6&60.4\\
         & CDAN~\cite{long2018conditional}&65.0&69.0&64.9&67.3&63.7&68.4&53.1&57.8&63.4&65.3&54.5&59.0&73.2&78.5&62.5&66.5\\
         & ENT~\cite{10.5555/2976040.2976107}& 65.2&71.0&65.9&69.2&65.4&71.1&54.6&60.0&59.7&62.1&52.1&61.1&75.0&78.6&62.6&67.6\\
         & MME~\cite{saito2019semisupervised}& 70.0&72.2&67.7&69.7&69.0&71.7&56.3&61.8&64.8&66.8&61.0&61.9&76.1&78.5&66.4&68.9\\
         &BiAT~\cite{ijcai2020-130}& 73.0& 74.9& 68.0& 68.8& 71.6& 74.6& 57.9& 61.5& 63.9& 67.5& 58.5& 62.1& 77.0& 78.6& 67.1& 69.7\\
         & APE~\cite{kim2020attract}& 70.4& 76.6& 70.8& 72.1 & 72.9& \textbf{76.7}& 56.7& 63.1& 64.5& 66.1& 63.0& 67.8& 76.6& 79.4& 67.6& 71.7\\
         & {\bf Pred\&Guide} & \textbf{74.7} & \textbf{79.2} & \textbf{75.9} & \textbf{74.8} & \textbf{74.8} & 75.9 & \textbf{66.0} & \textbf{69.5} & \textbf{71.4} &\textbf{71.2}& \textbf{68.7} & \textbf{72.5} & \textbf{78.5} & \textbf{80.2}&\textbf{72.9}&\textbf{74.8}\\
    \hline
    \end{tabular}}
    \caption{Semi-supervised DA performance ($\%$) on DomainNet dataset for both $1$ and $3$-shot protocols on 4 domains, R: Real, C: Clipart, P: Clipart, S: Sketch. 
    }
    \label{Visdatable}
\end{table*}
\section{Additional Analysis}
Here, we perform extensive analysis and ablation studies of the proposed Pred\&Guide framework. \\~\\
\textbf{Importance of Source Example Weighing}: 
To analyze the importance of weighting the source examples in the proposed Pred\&Guide framework, we run the algorithm with and without the source example weighing on the DomainNet dataset.
The results in Table \ref{tab:individual_components} show that when the algorithm is trained using the standard cross entropy loss $\Lagr_s$ without the weighting (i.e. {\em No weights}), the performance is considerably lower as compared to when the source samples are properly weighted.
Figure \ref{fig:near_far_examples} gives a visual illustration of the near and far source examples of the labeled targets for one class of DomainNet. 
We would like to highlight that the proposed source example weighting (SEW) can be seamlessly integrated with other SSDA methods to improve their performance. We experiment with a recent SSDA approach CDAC~\cite{CDAC}, and observe that SEW improves the mean performance ($\%$) of CDAC for both 1-shot and 3-shot settings from 73.6 to \textbf{74.4} and from 76.0 to \textbf{76.2} respectively.
\\~\\
{\bf Weighting Scheme: }
Now, we analyze the effectiveness of the proposed weighting strategy based on class-wise adaptability.
Specifically, we compare with 
fixed up-weighing and down-weighting.
For this experiment, we used fixed weights of $1.5$ and $0.5$ for the near and far samples respectively.
We observe from Table \ref{tab:weighing_comparison} that fixed weighting can give unpredictable results, i.e. for some cases (1-shot, R to C), it improves the results, and for other cases (3-shot R to S and R to C), the performance degrades. We observe that for almost all scenarios, the proposed source example weighting helps to boost the performance. 
We also compare with another popular weighting scheme, namely Focal Loss~\cite{lin2017focal}, which emphasizes on correct classification of hard training examples by weighing them more.
We observe that for all scenarios, the proposed technique outperforms focal loss.
As an example, for 3-shot setting, the proposed technique gives 72.5$\%$ and 79.2$\%$ for R to S and R to C as compared to 69.6$\%$ and 74.8$\%$ obtained using Focal loss.\\~\\

\begin{figure}[ht]
\centering
\includegraphics[width=0.2\textwidth]{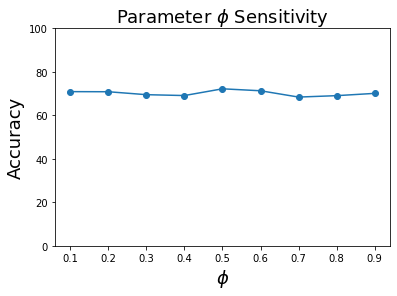}
\caption{Plot of hyperparameter $\phi$ vs. accuracy for 3 shot R to S setting of the DomainNet dataset}
\label{fig:sensitivity_analysis}
\end{figure}

\textbf{Near \& Far Weighing}: Next, we analyze the effect of weighing \textit{only} the near or far samples, and the results are reported in Table \ref{tab:individual_components}. 
Near source samples are defined as the samples having weight $w_i > 1$ and far samples are defined as the ones having $w_i < 1$. 
The results obtained using only UDA with self-training is shown in the first row.
If we introduce the labeled target samples right after this stage, as expected, the performance improves as shown in the second row. 
The third and fourth rows show the performance when only near and far away source examples are weighted. 
We observe that both these steps help to boost the performance.
The final row is the complete Pred\&Guide, which weighs both the near and far away source examples and outperforms all the other variants.
\\~\\
\textbf{Sensitivity Analysis of Hyperparameters}:
There are two primary hyperparamters in Pred\&Guide, $\phi$ which control the amount of weighing for the source examples and $T_n$. We plot the SSDA results (\%) for a wide range of $\phi$ values for the 3-shot R to S setting of DomainNet in Fig.~\ref{fig:sensitivity_analysis} (using Resnet34). 
We observe that the performance is quite stable for a wide range of parameter values, and even the lowest accuracy is better than the second best approach.
We also vary the values of $T_n$ between 1000 to 2000 for DomainNet and observe that the performance is stable for all the $T_n$ values. Note that we have used the same $\phi$ and $T_n$ values for all the domain settings of a dataset.

\begin{table}[ht!]
\centering
\scalebox{0.7}{
\begin{tabular}{|c|cc|cc|}
\hline
\multirow{2}{*}{Weighing Strategy}         & \multicolumn{2}{c|}{R to S} & \multicolumn{2}{c|}{R to C} \\
                                 & 1 shot       & 3 shot       & 1 shot       & 3 shot       \\ \hline
No Weights & \textbf{68.8} & 70.8 & 71.3 &     76.3 \\
Fixed Weights                    &     \textbf{68.8}        &    68.6          &  74.3            &     73.2         \\
Class-wise Adaptability Based Weights &      68.7        &        \textbf{ 72.5 }    &          \textbf{74.7}    &          \textbf{ 79.2}   \\ \hline
\end{tabular}
}
\caption{Performance of Pred\&Guide with no  weighting, fixed weighting and classwise adaptability based weighing of source examples.
}
\label{tab:weighing_comparison}
\end{table}
\begin{table}[ht]
\centering
\scalebox{0.7}{
\begin{tabular}{|cccc|cc|cc|}
\hline
\multicolumn{4}{|c|}{\multirow{2}{*}{Components}} & \multicolumn{2}{c|}{\multirow{2}{*}{R to S}} & \multicolumn{2}{c|}{\multirow{2}{*}{R to C}} \\
\multicolumn{4}{|c|}{} & \multicolumn{2}{c|}{} & \multicolumn{2}{c|}{} \\ \hline
UDA  & N  & F  & LT & 1 shot    & 3 shot    & 1 shot    & 3 shot    \\ \hline
 \checkmark& & & & 63.5&63.5 & 69.2 & 69.2 \\
 \checkmark& & & \checkmark & 68.8 & 70.8 & 71.3 & 76.3 \\
\checkmark& \checkmark&&\checkmark  & \textbf{69.0}  & 68.9  &  71.5  & 77.4\\
\checkmark&&\checkmark&\checkmark   &  68.1  &  71.9 & 72.4   & 77.3\\
\checkmark & \checkmark&\checkmark& \checkmark& 68.7& \textbf{72.5} & \textbf{74.7} & \textbf{79.2}\\
\hline
\end{tabular}
}
\caption{Ablation study for Pred\&Guide depicting performance of individual components using Resnet34: (1) UDA: Unsupervised domain adaptation with self-training, (2) N and F: class-wise source weighing of Near and Far examples respectively, (3) LT: labeled target examples are included in training starting at iteration $T_2$.}
\label{tab:individual_components}
\end{table}

\section{Conclusion} 
In this work, we addressed the importance of using labeled target samples effectively for the domain adaptation task in the semi-supervised setting. 
The proposed Pred\&Guide framework initially performs the domain adaptation in an unsupervised manner to avoid any bias that may be introduced due to the few labeled target examples. 
After this stage, we introduce two effective modules, labeled target prediction and source example weighting to effectively weigh the source examples to better guide the adaptation.
With Pred\&Guide, we set a new state-of-the-art for semi-supervised domain adaptation as illustrated by extensive experiments on two large-scale benchmark datasets. 

\ifCLASSOPTIONcaptionsoff
  \newpage
\fi



\bibliographystyle{IEEEtran}
\bibliography{IEEEabrv, main.bib}
\vspace{-1.6cm}
\begin{IEEEbiography}[{\includegraphics[width=0.8in,height=0.8in,clip, keepaspectratio]{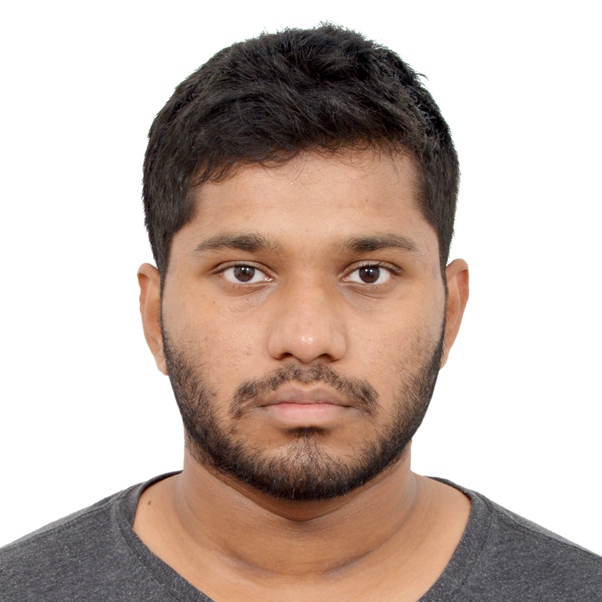}}]{Megh Bhalerao} received his Bachelor's of Technology Degree in Electrical Engineering from the National Institute of Technology Karanataka, India in 2020. He is currently an MS EE Student at the University of Washington Seattle.
\end{IEEEbiography}
\vskip -5\baselineskip  plus -1fil
\begin{IEEEbiography}[{\includegraphics[width=0.8in,height=0.8in,clip, keepaspectratio]{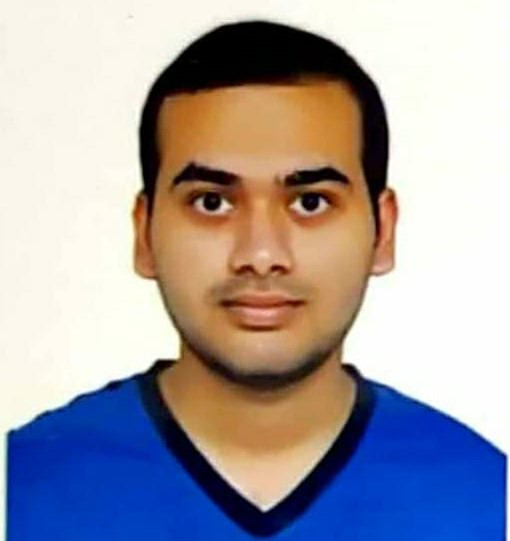}}]{Anurag Singh} 
is currently MS in Computer Science student at TU Munich. He completed his undergraduate in CS from NSIT, University of Delhi in 2018.
\end{IEEEbiography}
\vskip -4\baselineskip plus -1fil
\begin{IEEEbiography}[{\includegraphics[width=0.8in,height=1in,clip,keepaspectratio]{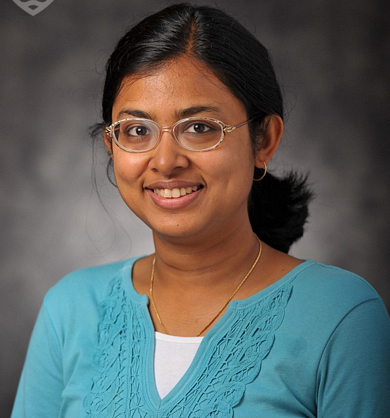}}]{Soma Biswas}
received the PhD degree in Electrical Engineering from University of Maryland, College Park in 2009. She is currently an Associate Professor in the Electrical Engineering Department, Indian Institute of Science, Bangalore. Her research interests is in Computer Vision, Pattern Recognition, Machine Learning and related areas. She is a Senior Member of IEEE.
\end{IEEEbiography}
%




%




\section{Supplimentary Material}
\subsection{Modularity of our Source Example Weighing Framework}
We can seamlessly integrate our source example weighing scheme with any given SSDA framework. In Table \ref{table_supp_cdac} we show that our proposed source example weighing framework can improve the performance of a recently proposed SSDA work Cross Domain Adaptive Clustering \cite{CDAC}. 
\begin{table*}[!htbp]

    \centering
    \scalebox{0.7}{
    \begin{tabular}{|c|c|c|c|c|c|c|c|c|c|c|c|c|c|c|c|c|c|c|}
    \hline
         \multirow{2}{*}{Net} & \multirow{2}{*}{Method} & \multicolumn{2}{c}{R to C} & \multicolumn{2}{c}{R to P} & \multicolumn{2}{c}{P to C} & \multicolumn{2}{c}{C to S} & \multicolumn{2}{c}{S to P} & \multicolumn{2}{c}{R to S} & \multicolumn{2}{c}{P to R} & \multicolumn{2}{|c|}{Mean}\\
         & & \multicolumn{1}{c}{1 shot} & \multicolumn{1}{c}{3 shot} & \multicolumn{1}{c}{1 shot} & \multicolumn{1}{c}{3 shot} & \multicolumn{1}{c}{1 shot} & \multicolumn{1}{c}{3 shot} & \multicolumn{1}{c}{1 shot} & \multicolumn{1}{c}{3 shot} & \multicolumn{1}{c}{1 shot} & \multicolumn{1}{c}{3 shot} & \multicolumn{1}{c}{1 shot} & \multicolumn{1}{c}{3 shot} & \multicolumn{1}{c}{1 shot} & \multicolumn{1}{c}{3 shot} & \multicolumn{1}{|c}{1 shot} &
         \multicolumn{1}{c|}{3 shot} \\
         \hline
         \multirow{2}{*}{ResNet34}
         &CDAC& 77.4 & 79.6 & 74.2 & 75.1 & 75.5 & 79.3 & 67.6 & 69.9 & 71.0 & 73.4 & 69.2 & 72.5 & 80.4 & 81.9 & 73.6 & 76.0\\
         & {\bf CDAC + SEW} & \textbf{78.1} & \textbf{80.1}  & \textbf{75.1}  & \textbf{75.5} & \textbf{75.7} & 78.7 & \textbf{68.7} & \textbf{70.7}  & \textbf{71.8} & \textbf{73.7} & \textbf{70.8} & \textbf{72.8} & \textbf{80.9} & \textbf{82.6} & \textbf{74.4} &  \textbf{76.2} \\
         \hline

    \hline
    \end{tabular}
    }
    \caption{Comparison of CDAC performance ($\%$) w.r.t. CDAC + SEW (source example weighing) on DomainNet dataset for both $1$ and $3$-shot protocols on 4 domains, R: Real, C: Clipart, P: Clipart, S: Sketch. 
    }
    \label{table_supp_cdac}
\end{table*}

\subsection{Comparison of SEW with standard focal loss}
In table \ref{focal_supp}, we compare our proposed source example weighing scheme with the standard focal loss which assigns weights to examples based on hardness. Our SEW scheme is in contrast to focal loss where we assign example weights based on domain similarity.
		\begin{table*}[t]
    \centering
    \begin{tabular}{|c|c|c|}
    \hline
	Weighting-Strategy  &  R to S	& R to C	\\
	\hline
	No Weights & 70.8 & 76.3 \\
	Fixed Weights \color{black} & 68.6 & 73.2 \\
	Focal Loss 	&69.6	&74.8 \\	
	Proposed Class-wise Adaptability Based Weights &72.5	&79.2\\	
	\hline
	\end{tabular}
    \caption{Comparison of the proposed weighting scheme and Focal loss in 3-Shot setting on Domainnet dataset with ResNet-34 as backbone.}
    \label{focal_supp}
    \end{table*}
\end{document}